\documentclass[conference]{IEEEtran}
\IEEEoverridecommandlockouts
\usepackage{cite}
\usepackage{amsmath,amssymb,amsfonts}
\usepackage{algorithmic}
\usepackage{graphicx}
\usepackage{textcomp}
\usepackage{xcolor}
\usepackage{hyperref}
\def\BibTeX{{\rm B\kern-.05em{\sc i\kern-.025em b}\kern-.08em
    T\kern-.1667em\lower.7ex\hbox{E}\kern-.125emX}}

\DeclareRobustCommand*{\IEEEauthorrefmark}[1]{%
  \raisebox{0pt}[0pt][0pt]{\textsuperscript{\footnotesize #1}}%
}

\usepackage{fancyhdr}
\begin{document}

\title{On the use of Vision-Language models for Visual Sentiment Analysis: a study on CLIP
}

\author{\IEEEauthorblockN{Cristina Bustos\IEEEauthorrefmark{1}, Carles Civit\IEEEauthorrefmark{1}, Brian Du\IEEEauthorrefmark{1}, Albert Sol\'e-Ribalta\IEEEauthorrefmark{1} and Agata Lapedriza\IEEEauthorrefmark{1,2}}

\IEEEauthorblockA{\IEEEauthorrefmark{1} Universitat Oberta de Catalunya, Barcelona, Spain
%\{mbustosro,xbaro,alapedriza\}@uoc.edu
}
\IEEEauthorblockA{\IEEEauthorrefmark{2} Northeastern University, Boston, MA, USA}
Email: \{mbustosro, ccivitsa, gdu, asolerib, alapedriza\}@uoc.edu
}

\maketitle
\thispagestyle{fancy}
\maketitle

\begin{abstract}
This work presents a study on how to exploit the CLIP embedding space to perform Visual Sentiment Analysis. We experiment with two architectures built on top of the CLIP embedding space, which we denote by CLIP-E. We train the CLIP-E models with WEBEmo, the largest publicly available and manually labeled benchmark for Visual Sentiment Analysis, and perform two sets of experiments. First, we test on WEBEmo and compare the CLIP-E architectures with state-of-the-art (SOTA) models and with CLIP Zero-Shot. Second, we perform cross dataset evaluation, and test the CLIP-E architectures trained with WEBEmo on other Visual Sentiment Analysis benchmarks. Our results show that the CLIP-E approaches outperform SOTA models in WEBEmo fine grained categorization, and they also generalize better when tested on datasets that have not been seen during training. Interestingly, we observed that for the FI dataset, CLIP Zero-Shot produces better accuracies than SOTA models and CLIP-E trained on WEBEmo. These results motivate several questions that we discuss in this paper, such as how we should design new benchmarks and evaluate Visual Sentiment Analysis, and whether we should keep designing tailored Deep Learning models for Visual Sentiment Analysis or focus our efforts on better using the knowledge encoded in large vision-language models such as CLIP for this task. Our code is available at \url{https://github.com/cristinabustos16/CLIP-E}.
\end{abstract}

\begin{IEEEkeywords}
visual sentiment analysis, vision-language models, zero-shot classification
\end{IEEEkeywords}

\section{Introduction}

% interest of visual sentiment analysis
Visual Sentiment Analysis studies how to automatically recognize affect in visual content. The interest for recognizing affect in images has significantly increased in the past few years, due to the popularity of social media and the multiple applications of this topic in areas like opinion mining \cite{ye2019visual}, affective image retrieval \cite{pang2015deep}, education \cite{lopez2017mining}, mental health \cite{maity2022multitask}, %guntuku2019twitter, ramirez2020detection}, 
or hate speech analysis \cite{patwa2021findings}. Additionally, feature representations of image sentiment have been shown to be meaningful for emotion recognition in scene context \cite{kosti2019context}.

\begin{figure}[h]
\centering
\includegraphics[width=0.48\textwidth]{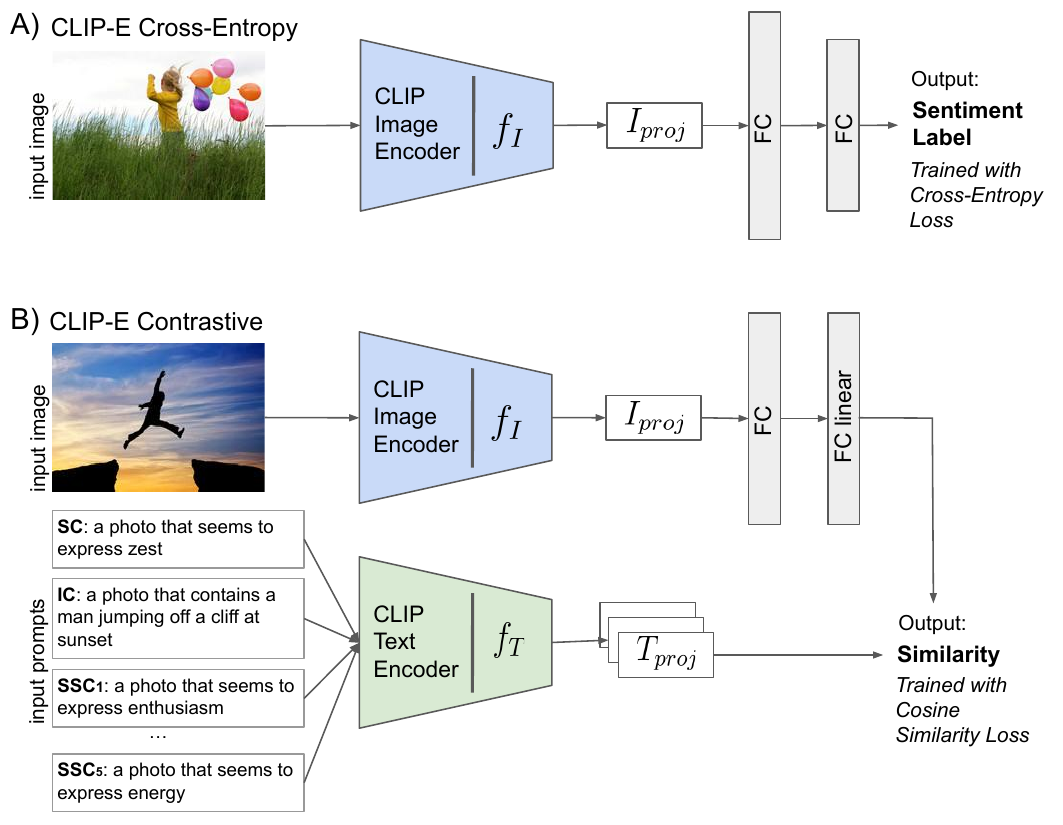}
\caption{CLIP-E architectures. (A) CLIP-E with Cross-Entropy loss; (B) CLIP-E with Contrastive loss. In these architectures the CLIP Image Encoder and the CLIP Text Encoder are frozen.}
\label{fig:architecture}
\end{figure}

% what do we do in this work: first contribution
In this work we study how to use the CLIP embedding space for Visual Sentiment Analysis. We experiment with two simple Deep Learning (DL) architectures built on top of the CLIP embedding space with two different loss functions: Cross-Entropy loss and Contrastive loss. We compare the performance of these CLIP-based approaches (denoted by CLIP-E, from CLIP Emotions) with state-of-the-art (SOTA) models and Zero-Shot CLIP. For our experiments we use the WEBEmo dataset \cite{panda2018contemplating}, which is the largest public and manually annotated visual sentiment analysis benchmark. Interestingly, we observe that the CLIP-E approaches obtain competitive results when compared with SOTA models on the binary valence problem (positive vs. negative emotion), while they significantly outperform SOTA models in the finest grained emotion classification. Additionally, CLIP-E with Contrastive loss provides a flexible model for Visual Sentiment Analysis, able to deal with different emotion taxonomies. The CLIP-E architectures are illustrated in Fig. \ref{fig:architecture} and explained in Sect.\ref{sec:clipe}.

%\agata{mention that clip-e is computationally much less expensive than other tailored approaches} This means that we can obtain competitive results on visual sentiment analysis with the CLIP-

% explain previous works that explore the knowledge of CLIP on emotions
Our motivation to build on top of the CLIP embedding space for Visual Sentiment Analysis is supported by several works showing that CLIP encodes affect-related information. In particular, some works have reported high accuracy results for CLIP Zero-Shot classification on Facial Emotion Recognition \cite{CLIP} or Visual Sentiment Analysis \cite{bondielli2021leveraging}, while others show emotional concepts, such as happiness or surprise, are encoded in the the CLIP abstract visual features \cite{goh2021multimodal}. %(more details on CLIP and these works are provided in Sect.\ref{sec:related_work}). 
Our results support the idea that the rich visual-semantic representation learned by the CLIP embedding space already contains meaningful general knowledge for effective Visual Sentiment Analysis.
%Furthermore, the CLIP embedding space constitutes a rich representation for both visual features and semantics, including commonsense knowledge. Visual Sentiment Analysis is a high level task that requires a connection between the visual features and the affect represented in the image or evoked by the image, and some of these connections already encoded in the CLIP embedding space.

%First, % different works show that large vision-langauge models such as CLIP perform surprisingly well on recognition tasks with zero-shot classification. In particular, 
%some works provide high-accuracy results on CLIP zero-shot classification for Facial Emotion Recognition \cite{CLIP} or Visual Sentiment Analysis \cite{bondielli2021leveraging}, while other works show that CLIP abstract visual features encode emotional concepts such as happiness or surprise \cite{goh2021multimodal}. This means that the CLIP embedding space already contains exploitable affect-related knowledge. Second, the CLIP embedding space constitutes a rich representation for both visual features and semantics, including commonsense knowledge. Visual Sentiment Analysis is a high level task that requires a connection between the visual features and the affect represented in the image or evoked by the image, and some of these connections already encoded in the CLIP embedding space.

% what do we do in this work: second contribution
In this work we also study the trade-off between specialization and generalization  through cross-dataset evaluation. We train the CLIP-E approaches with WEBEmo and then we test them on other popular visual sentiment benchmarks. We observe that models obtaining the highest accuracies on the WEBEmo test set, perform more poorly when tested in other datasets (for example, the CLIP-E approaches and CLIP Zero-Shot outperform the curriculum learning approach from Panda et al. \cite{panda2018contemplating}, when it is learned on WEBEmo and tested in Emotion-6 binary benchmark). Furthermore, the CLIP-E approaches show an interesting balance between specialization and generalization (for instance, CLIP-E Cross-Entropy achieves SOTA results on WEBEmo 6 category and 25 category benchmarks, while it also achieves the best or second best cross-dataset results in 4 out of 5 benchmarks). Surprisingly, CLIP Zero-Shot outperforms all the models trained on WEBEmo when they are tested on the FI \cite{you2016building} dataset (the second largest manually annotated public benchmark for image sentiment analysis).

% discussion
Our results motivate interesting observations and questions that we discuss in Sect.\ref{sec:discussion}. For example, we observe that often the results provided by different methods correspond to reasonable different interpretations of the same input. This motivates questioning the common multi-class approach for Visual Sentiment Analysis, which assumes that there is just one acceptable ground truth category for each input. We suggest that future efforts for collecting data should consider collecting multiple labels per each sample, and approach the visual Sentiment Analysis problem from a multi-label perspective. Also, we observe that customized Deep Learning architectures specifically designed and trained for Visual Sentiment Analysis might have less generalization capacity, particularly when compared to models that build on top of the knowledge encoded in large vision-language models. This motivates further explorations on how to leverage knowledge from vision-language models for Visual Sentiment Analysis and, more generally, for other visual affect recognition tasks. 
%specialized in achieving the best accuracy for a single dataset? We observe that customized and complicated architecture may have the benchmark for a dataset, but they lack in generalization when tested in other type of data.} \agata{Cristina, complete this after the current questions we post in the discussion. Keep it in color, so that I can easily review afterwards}.

% summary of contributions, code available, etc.
In summary, we empirically show that it is possible to leverage the CLIP embedding space for visual sentiment analysis. In particular, the CLIP-E simple architectures, trained on top of the CLIP embedding space, are able to outperform models that are explicitly designed for the task of Visual Sentiment Analysis in fine grained classification benchmarks. Additionally, the CLIP-E architecture with Contrastive loss results in a flexible model for visual sentiment analysis that can deal with the different emotion taxonomies used in different datasets. Finally, the affect knowledge embedded in the CLIP space allows the CLIP-E architectures to generalize better, as shown in our cross-dataset evaluations. %All the code and material generated with the experiments will be publicly released for academic research purposes upon the acceptance of the paper. 

\section{Related work}
\label{sec:related_work}

In general, Visual Sentiment Analysis is approached with supervised learning. The first studies were conducted on small datasets, and the proposed methods were based on extracting hand-crafted features from the images \cite{yanulevskaya2008emotional, machajdik2010affective, zhao2014exploring}. 
%They were either low level features (color or texture descriptors), or higher level features designed from psychology or art perception theories. 
Later, some works focused on
%attempted to bridge the gap between the pixel-level representation and the affective content of the image by 
automatically detecting a large set of mid-level concepts or attributes in the images, and then training a machine learning model on top of this mid-level representation \cite{borth2013sentibank}. 
After that, the availability of larger-scale affective datasets \cite{panda2018contemplating, you2016building, katsurai2016image, zhao2021affective} allowed the exploration of end-to-end DL methods, and multiple works showed the superior performance of DL methods with respect to the use of hand-crafted features \cite{chen2014deepsentibank,campos2017pixels,zhao2021affective}.
%,yang2018weakly}. 
Researchers have been exploring different approaches to improve these end-to-end DL models. Some works study the use of attention mechanisms to enforce the model to focus on specific regions of the image \cite{fan2017role}. Based on this idea, Zhang and Xu \cite{zhang2020weakly} proposed a model that integrates a prediction of emotion intensity maps during the learning process. Concretely, the network has a first classification stream, followed by an emotion intensity prediction stream that uses Class Activation Map \cite{zhou2016learning} to guide the emotion intensity learning. Then, the predicted intensity map is integrated into the final classification stream. At the same time, Panda et al.\cite{panda2018contemplating} explored the advantages of curriculum learning, empirically showed the effectiveness of a guided training strategy that exploits the hierarchical structure of emotions by learning the easiest task first (binary classification, positive vs. negative emotion), and then gradually learning to recognize finer grained emotion categories. Also, other interesting recent works explore the use of image and emoji pairs obtained from social networks as weak labels to train the Visual Sentiment Analysis models \cite{al2019smile,tsimpoukelli2021multimodal}. While the accuracies are quite remarkable, the results are not as good as the ones obtained with manually labelled images. 

Some works have explored the use of external knowledge and semantic representations to bridge the gap between the pixel level representations and the affective content. For example, Zhan et al.\cite{zhan2019zero} created an affective embedding space using Sentibank's ANPs and visual features in parallel, which also allowed them to perform zero-shot sentiment analysis. Later, Wei et al. \cite{wei2020learning} introduced a set of 690 emotion words and searched the web for 1M related images to build the StockEmotion dataset. With it, the authors built a multi-modal pipeline with 3 inputs: emotion words associated with the image searched, the image, and other non-emotion text adjacent to the image when it was found. They trained the model and segregated the image feature extraction portion of the network. When testing against other datasets, they integrated the image feature extractor into a classifier and fine tuned it to the task. Later, Ortis et al.\cite{ortis2021exploiting} used a two branch network for the semantic and visual streams. They extracted text detections from the image, through the use of four image descriptor models and post-process the resulting texts into a BoW weighted with the predominant score. In parallel, they use a pretrained CNN trained on Sentibank to extract the feature layer. With this, they apply an SVM classifier to both outputs to obtain the resulting emotion.

More recently, Xu et al. presented MDAN \cite{xu2022mdan}, a multi-level attention network that exploits a similar concept by incorporating the insight of the hierarchical relationship among emotions at different affective levels into the model. More concretely, the model has two branches: a bottom-up branch that focuses on avoiding hierarchy violation (an image cannot be \emph{happy} (fine-grained emotion) and \emph{negative} (binary)); and a top-down branch that focuses on learning emotions at a high level, thus benefiting the learning of finer-grained emotions (it is easier to infer that an image is \emph{happy}, if it is first known that it is \emph{positive}). Currently, MDAN is the method that obtains the state-of-the-art results on WEBEmo.

\subsection{CLIP for Affect Recognition Tasks}
CLIP \cite{CLIP} is a popular and widely used large Vision-Language model, trained on millions of $image-text$ pairs (where $text$ are image captions) using a contrastive learning approach. In short, CLIP has an image embedding and a text embedding, and the model is trained to embed $image$ and $text$ that belongs together into similar points (according cosine similarity), while pushing away the embeddings of $image$ and $text$ pairs that do not belong together. Then, CLIP can be used to perform zero-shot image classification by projecting images and text prompts in the embedding space, and classifying the images with the prompt whose embedding has the highest similarity with respect to the embedded image.
%The learning objective of CLIP's training was to identify the most relevant text description for a given image. By leveraging natural language supervision, CLIP is capable of identifying natural language concepts embedded within images.

Zero-Shot classification with CLIP performs remarkably well in a large variety of tasks \cite{CLIP}. In particular, a few works explicitly study the capacity of CLIP Zero-Shot for affect-related tasks. Actually, the authors of CLIP already showed very interesting results on Facial Emotion Recognition. Also, Bodinelly et al. \cite{bondielli2021leveraging} reported remarkable results with zero-shot CLIP and fine-tuned CLIP on Image-Emotion \cite{you2016building}, a collection of 3+ million images, weakly labeled on 8 emotion categories. More recently, Deng et al. \cite{deng2022simple} presented SimEmotion, a model that combines the CLIP vision and language features for Visual Sentiment Analysis. The architecture combines cosine similarity (to supervise $image$ and $text$ pairs) and cross-entropy loss (to supervise the image emotion categorization). In their case the whole image embedding is fine tuned. Their approach also leverages the CLIP embedding space for Visual Sentiment Analysis, but it requires much more training than the CLIP-E approaches. Furthermore, SimEmotion relies on Cross-Entropy to classify new unseen images, meaning that the model can not be tested on taxonomies that differ from the taxonomy used during training (for example, if the model is trained on a taxonomy with $6$ emotion categories, the model can not be directly used for testing on a finer grained categorization with more than $6$ emotion categories). Despite this, SimEmotion is a very interesting architecture. Unfortunately, the authors did not release their code, models, and neither the data partitions used in their experiments, which makes it impossible to compare our computationally less expensive CLIP-E approaches with SimEmotion.

\section{Visual sentiment analysis with CLIP-E}
\label{sec:clipe}

% In this study we use the large and popular Vision-Language model CLIP \cite{CLIP}, due to its demonstrated potential for generalizing across different domains like image classification, object detection and natural language processing. CLIP was trained on a large dataset composed by image-text pairs using a contrastive learning approach, where the model is trained to predict whether an image and its associated text belong together or not. The learning objective of CLIP's training was to identify the most relevant text description for a given image. By leveraging natural language supervision, CLIP is capable of identifying natural language concepts embedded within images. 

Our goal is to leverage on the knowledge that CLIP has learned to perform Visual Sentiment Analysis with minimal training effort. For that we train two simple DL architectures, denoted by CLIP-E, on top of the CLIP embedding space. One of the  CLIP-E architectures is trained with Cross-Entropy loss (CLIP-E Cross-Entropy) and the other is trained with Contrastive loss (CLIP-E contrastive). The architectures are illustrated in Fig. \ref{fig:architecture}.

\textbf{CLIP-E Cross-Entropy}. For the CLIP-E Cross-Entropy (Fig.\ref{fig:architecture}.A), we just use the CLIP image encoder with two additional fully connected layers, and then we use cross-entropy loss on emotion categories to train the two additional fully connected layers (i.e. the CLIP image encoder is frozen). 

\textbf{CLIP-E Contrastive}. For the CLIP-E Contrastive (Fig.\ref{fig:architecture}.B), we add two fully connected layers on top of the image encoder to allow deformations of the embedding space, and then we use Contrastive loss with cosine similarity with a collection of text prompts. During training, both the CLIP image and text encoders are frozen. For the CLIP-E Contrastive approach, we generated different image caption types for an image $I$: $SC$, which refers to sentiment caption; $SSC$, which refers to sentiment synonyms captions; and $IC$, which refers to image caption. Concretely, given a Image $I$ with their respective sentiment class $C$, we use for $SC$ the same sentiment prompting pattern as in \cite{deng2022simple}. Thus, $SC$ is the prompt "a photo that seems to express $C$" (e.g "a photo that seems to express contentment"). For $SCC$ we used the language generation model ChatGPT \cite{chatgpt} to obtain the 5 most significant synonyms of class $C$ ($Csyn_1,...,Csyn_5$)\footnote{The complete list of synonyms is provided in the supplementary material}. Then, $SSC$ is the collection of prompts "a photo that seems to express $Csyn_i$", for i = 1,...,5. Finally, for $IC$ we use the image captioning from OFA image caption \cite{wang2022unifying} to obtain a prompt that is a short description of the image. Fig. \ref{fig:architecture}.B shows the $SC$, $IC$, and $SSC_m$ prompts for the displayed input image. 
%CLIP model has two encoder architectures: Visual Encoder $f_{I}$ and Text encoder $f_{T}$. Both encoders project on the CLIP embedding space. 

%With the CLIP encoders, we tested two methodologies, one using cross entropy loss for image sentiment classification and the other, using contrastive loss in order to find the similiraties between sentiment captions and images.

\subsection{Training and inference details for CLIP-E}

\textbf{CLIP-E Cross-Entropy}. For the CLIP-E Cross-Entropy architecture we add, on top of the CLIP image embedding, one fully connected layer of 512 units followed by the prediction layer. These two layers are trained with the well known Cross-Entropy Loss (the CLIP image embedding is frozen). In this case the inference for an unseen image $I$ is straightforward: the image is classified as the class with highest probability. Notice that inference with CLIP-E Cross-Entropy can be only made with the same taxonomy used for training.

\textbf{CLIP-E Contrastive}. For the CLIP-E Contrastive architecture we added a fully connected layer on top of the CLIP image encoder with 512 units and relu activation, and then a fully connected with a linear activation function with 512 units before the loss function. The linear activation function was chosen in order to not lose the negative activations, because the original CLIP trained embedding space has both negative and positive values. To train the model we use the same contrastive loss function used in the original CLIP approach \cite{CLIP}. Concretely, given a batch of N $image$-$text$ pairs ({$(I_i,T_i)$}, $i=1,...,N$), and then we compute $L_{img}$ and $L_{text}$ as follows

\begin{equation}
    L_{img} = \frac{1}{N}\sum_{i=1}^{N}\Big[ -\log\frac{\exp(\langle I_{ei}, T_{ei} \rangle)}{\sum_{j=1}^{N} \exp( \langle I_{ei}, T_{ej}\rangle )} \Big]
\end{equation}

\begin{equation}
L_{text} = \frac{1}{N}\sum_{i=1}^{N}\Big[ -\log\frac{\exp(\langle T_{ei}, I_{ei} \rangle )}{\sum_{j=1}^{N} \exp( \langle T_{ei}, I_{ej}\rangle)} \Big]
\end{equation}

where $\langle I_{ei}, T_{ej} \rangle$ is the cosine similarity between the embedding vectors of the $i$-th image sample ($I_{ei}$) and the $j$-th text sample ($T_{ej}$), and $\langle I_{ei}, T_{ei} \rangle$ is the cosine similarity between the $i$-th image sample and its corresponding text caption. Then, the Contrastive Loss ($L_{CO}$) is computed by

\begin{equation}
L_{CO} =  \frac{L_{Img} + L_{text}}{2}
\end{equation}
This loss encourages the model to maximize the similarity between the embedding vector of each sample and its text description, and minimize the similarity between the embedding vectors of each sample and the other samples in the batch.

Once that CLIP-E Contrastive is trained, we proceed as follows for the inference on an unseen image $I$. We create a list of image sentiment captions $SC={SC_{1},SC_{2} ... SC_{M}}$, where $M$ is the number of  sentiment classes in the supervised vision approach (including synonyms). Then, we compute the cosine similarity between the embedded image $I_e$ and each of the embedded sentiment captions $T_{SC_m}$, 

\begin{equation}
\langle I_{e}, T_{SC_m}\rangle =  \frac{I_{e} \cdot T_{SC_m}}{\| I_{e}\| \| T_{SC_m}\|}
\label{eq:cosine}
\end{equation}

and then we compute $\hat{m}$ by 

\begin{equation}
\hat{m} = arg\_max_{m=1,...,M} (\langle I_{e}, T_{SC_m} \rangle )
\label{eq:max_cosine}
\end{equation}

Finally, image $I$ is then classified as the class of the taxonomy that corresponds to the prompt $SC_{\hat{m}}$. For example, let us suppose we are computing inference for an image using the 25 category taxonomy of WEBEmo. Then, imagine that the most similar prompt is 'a photo that seems to express positivity'. In this case the corresponding class for this prompt would be \textit{positivity}, which is one of the 5 synonyms for the category \textit{optimism}. Therefore, the image would be classified as \textit{optimism}. Also, if we want to classify the image in binary valence, we would could use the WEBEmo taxonomy, and classify the image as \textit{positive}, since \textit{optimism} belongs to the \textit{positive} valence cluster. Notice that inference with CLIP-E Contrastive can be made with any same taxonomy, since we can use an arbitrary set of $SC$ prompts during inference.

Both CLIP-E models were trained until 15 epochs. The learning rate was initialized at 1e-3, and it is scheduled to decrease by a factor of 0,1 when the validation loss was in a plateau. We used Adam optimizer, and the training is stopped once the validation loss does not increase any longer. The batch size for CLIP-E Cross-entropy was 32 and for CLIP-E Contrastive was 8. %Our code will be made publicly available for reproducibility. 

\section{Experiments}

We perform two sets of experiments. First, we train and test the CLIP-E approaches on WEBEmo \cite{panda2018contemplating}, the largest publicly available dataset with manual annotations on visual sentiment. We compare the obtained results with SOTA models and with CLIP Zero-Shot. Second, we perform cross-dataset evaluation. Concretely, we train different models on WEBEmo and test on IAPS \cite{lang1997international}, Emotion-6 \cite{panda2018contemplating} (2 and 6 category benchmarks), EmotionROI \cite{peng2016emotions} (2 and 6 category benchmarks), and FI \cite{you2016building} (2 and 8 category benchmarks). The next subsection provides a short description of the datasets used in our experiments. 

\subsection{Datasets}

In this work, we use WEBEmo\cite{panda2018contemplating} to train the models. WEBEmo is the largest manually annotated dataset for visual sentiment analysis and a popular benchmark. The dataset contains 268,000 images retrieved from the Internet. The images are labeled with the Parrott’s hierarchical model with primary, secondary and tertiary emotions (2 emotion polarity categories -positive/negative-, branch out to 6 emotion categories, which, in turn, expanded into 25 fine-grained emotion categories). The dataset was collected by crawling through the 25 emotions as keywords over the internet, while duplicates and non-english tagged images were removed.  
%Examples of this dataset is shown in Fig. \ref{fig:example_WEBEmo}. 

% \begin{figure}[h]
% \centering
% \includegraphics[width=0.50\textwidth]{figures/example_images_WEBEmo.png}
% \caption{Randomly selected images from WEBEmo dataset. \agata{Esta imagen se ve un poco pobre. Quizás habría que poner ejemplos de las 6 categorías, una categoría por columna}.}
% \label{fig:example_WEBEmo}
% \end{figure}

In our cross-dataset experiments we use 4 different smaller scale datasets for evaluating the models, all of them manually labeled. The FI \cite{you2016building} is a collection of 23,308  images from Flickr and Instagram manually labeled on 8 emotion categories by Amazon Mechanical Turk workers using the sentiment taxonomy of Mikel's emotions \cite{mikels2005emotional}. These emotions are also grouped in two valence categories (positive vs. negative). The IAPS dataset \cite{lang1997international} is composed of 1,282 images, labeled with binary valence. The  Emotion-6 dataset \cite{panda2018contemplating} contains 8,350 images, and is labelled with 6 emotion categories. The categories are also grouped in two valence categories. Finally, the EmotionROI \cite{peng2016emotions} is composed of 1,980 images, and has binary valence labels as well as 6 category labels.

\subsection{Results on WEBEmo}

We trained the CLIP-E models on WEBEmo using the training and validation data partitions published by the authors of WEBEmo. The models are tested in the test partition, composed of approximately 53K images. Table \ref{tab:WEBEmo_accuracy} summarizes the results on WEBEmo test set. The table includes $random$ and \emph{most represented class} as reference results. Then it includes three SOTA models (Panda et al. \cite{panda2018contemplating}, Zhang et al. \cite{zhang2020weakly} and MDAN \cite{xu2022mdan}), CLIP Zero-Shot, and the CLIP-E approaches. We run our models using 5 different seeds and the average standard deviation is 0,24 on a 0-100 scale.
%The CLIP Zero Shot refers to the results made using the knowledge that the publicly available model of CLIP already knows. CLIP Zero Shot shows the potential in language-vision models on affective recognition tasks, the zero shot accuracy surpassed the accuracy of the baseline models like random or most represent class, this indicates that in the embedding space of vision and semantic there is already some knowledge about sentiment.

\begin{table}[htbp]
\caption{WEBEmo Classification accuracies. Comparison of CLIP-E approaches with CLIP Zero-Shot and SOTA models and Ablation Study on CLIP-E Contrastive.}
\begin{center}
\begin{tabular}{|l|c|c|c|}
\hline 
\textbf{Model} & \textbf{2 cat}& \textbf{6 cat}& \textbf{25 cat} \\
\hline
\textit{Random}  & 50.00 & 16.66  & 4.00  \\
\hline
\textit{Most represented class} & 54.31 & 29.01 & 11.38 \\
\hline
Panda et al. \cite{panda2018contemplating} & 81.41* & -- &  -- \\
\hline
Zhang et al. \cite{zhang2020weakly} & 78.17 & 52.32 & 32.14 \\
\hline
MDAN \cite{xu2022mdan}  & \textbf{82.72} & 55.65  & 34.28 \\
\hline

CLIP Zero Shot & 72.45 & 38.96 &  11.25 \\
\hline
CLIP-E Cross-Entropy & 80.02 & \textbf{56.42} & 	\textbf{40.65}  \\
\hline
CLIP-E Contrastive & 79.68	& 55.77 * & 34.68 * 	 \\
\hline
\hline
\textbf{CLIP-E Contrastive Ablation} & & &  \\
\hline
CLIP-E Contrastive [SC] &    77.62 &	54.39	& 33.72  \\
\hline
CLIP-E Contrastive [IC] & 64.40 &	38.59 &	11.42   \\
\hline
CLIP-E Contrastive [SC, IC] & 78.59 & 55.67 & 34.44 \\
\hline
CLIP-E Contrastive [SC, SSC] & 79.02 & 53.79 & 31.89  \\
\hline

\end{tabular}
\label{tab:WEBEmo_accuracy}
\end{center}
\end{table}

We observe that both CLIP-E models achieve state-of-art accuracy on the finegrained categories (6 category and 25 category benchmarks). Also, in binary classification the accuracy obtained by the CLIP-E models is equivalent and competitive with the SOTA counterparts MDAN  \cite{xu2022mdan} and Zhang et al. \cite{zhang2020weakly}. The advantage of our models is that we do minimal training effort using simple deep learning architectures built on top of the frozen CLIP embeddings. 

\begin{figure}[t]
\centering
\includegraphics[width=0.49\textwidth]{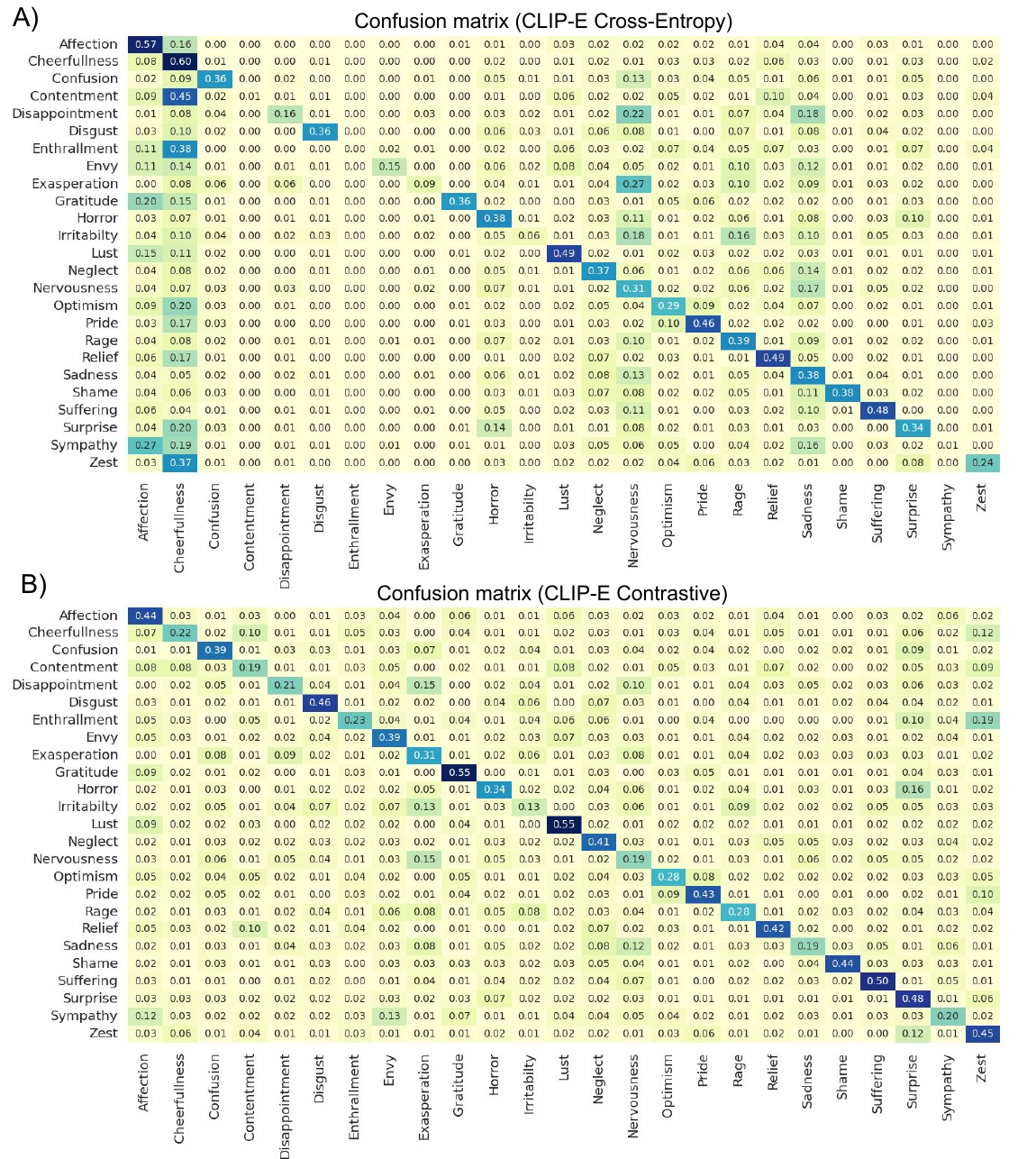}
\caption{Confusion Matrices for CLIP-E Cross-Entropy (A) and CLIP-E Contrastive (B) when trained and test on WEBEmo.}
\label{fig:cm}
\end{figure}

% We leverage in the semantic and visual information pairing learn by the CLIP. 
The confusion matrix obtained with each of the CLIP-E models are depicted in Fig.\ref{fig:cm}. We notice that CLIP-E Cross-Entropy achieves higher accuracies, but the confusion matrix shows more errors on underrepresented classes, such as \textit{sympathy, exasperation}, or \textit{disappointment}. We also observe confusions on classes that are semantically close, such as \textit{contentment} and \textit{cheerfulness}. In contrast, we see less of these errors in  the CLIP-E Contrastive confusion matrix. 
%CLIP-E Contrastive can generalize better with classes underrepresented and classification more finegrained between similar sentiment classes. 

In fine-grained classification, both CLIP-E models show higher performance in comparison with Zhang et al. \cite{zhang2020weakly}. Particularly, in categories like 'zest', 'shame', 'relief' CLIP-E Contrastive obtains accuracies of 45\%, 44\% and 42\% in those classes, respectively, while Zhang et al. \cite{zhang2020weakly} obtained 10\%, 0\% and 0\% in the same classes. 
%We can not compare fine grained accuracies with MDAN \cite{xu2022mdan} because authors have not shared the confusion matrix.

\textbf{Ablation results for CLIP-E Contrastive}. Table \ref{tab:WEBEmo_accuracy} also shows ablation results for CLIP-E Contrastive (last 4 rows). Concretely, we tested the model using separately the different types of prompts. We trained the model only using SC (sentiment caption), only using the IC (image caption), using IC and SC, and SC with SSC (with the synonyms sentiment caption). We observe that just using the sentiment caption with synonyms, the CLIP-E Contrastive achieves competitive accuracies when compared with the full model and with SOTA. However, training only with IC does not produce good results and performs worse than CLIP Zero-Shot. This happens because the model is not trained with the goal of learning about sentiment, it is been trained for learning to correctly pair images and their respective caption descriptions. However, the IC information is complementary when it is combined with SC, achieving higher accuracy than just training solely with SC. When SC and SSC are combined the performance on fine grained classification drops a bit, while the performance on binary classification improves. Overall, the best accuracy is achieved by the full CLIP-E Contrastive final model, when the information from SC, IC and SSC is combined all together. 

%\begin{table}[htbp]
%\caption{CLIP Contrastive finetuning ablation studies }
%\begin{center}
%\begin{tabular}{|c|c|c|c|}
%\hline 
%\textbf{Text Configuration} & \textbf{2 cat}& \textbf{6 cat}& \textbf{25 cat} \\
%\hline
%SC &    77,45 &	53,63	& 33,72  \\
%\hline
%IC & 64,40 &	36,59 &	11,42   \\
%\hline
%SC, IC & 78,42 & 54,85 & 34,44 \\
%\hline
%SC, SSC & 79,02 & 53,00 & 31,89  \\
%\hline
%SC, SSC, IC & 79,57  &  55,50  & 34,68  \\
%\hline

%$\end{tabular}
%\label{tab1}
%\end{center}
%\end{table}

\subsection{Cross-dataset evaluation}

\begin{table*}[htbp]
\caption{Results of the cross-dataset evaluation. The models that require training are trained on WEBEmo and tested on the different Visual Sentiment Analysis benchmarks. Bottom part provides the ablation study on CLIP-E Contrastive.}
\begin{center}
\begin{tabular}{|l|c|c|c|c|c|c|c|}
\hline 
\textbf{Model} & IAPS &  Emotion-6 2 Cat &  Emotion-6 6 Cat & EmotionROI 2 Cat & EmotionROI 6 Cat & FI 2 Cat &  FI 8 Cat \\
\hline
\textit{Random} & 50.00 & 50.00 & 16.66 & 50.00 & 16.66 & 50.00 & 12.50 \\
\hline
\textit{Most represented class} &57.76 & 61.07 & 26.56 & 66.66 & 16.66 & 70.31  & 23.09 \\
\hline
Panda et al. \cite{panda2018contemplating} & --	& 78.38& 	X	& --& X & 71.42& --\\
\hline
CLIP Zero-shot &  74.37&	78.50	&46.53	&77.87	&42.17	& \textbf{84.5}	& \textbf{49.97}\\
\hline
CLIP-E Cross-Entropy & \textbf{85.32}	& \textbf{84.56} & 	(\textbf{58.39})	& \textbf{85.67}	& X & 	83.06*& 	X\\
\hline
CLIP-E Contrastive & 81.87 & 81.01 & 51.71 & 80.75* & 43.18* & 79.23 & 38.07* \\
\hline
\hline\textbf{CLIP-E Contrastive Ablation} & & & & & & &\\
\hline 
CLIP-E Contrastive [SC] & 78.75 & 	77.55& 	50.62	& 71.71	& 36.72	& 59.6	& 20.08 \\
\hline
CLIP-E Contrastive [IC] & 63.75 & 	69.34	 & 40.79 & 	71.26 &  28.43	 & 79.19	 & 35.89\\
\hline
CLIP-E Contrastive [SC, IC] & 76.25	 & 80.42 & 	52.58 & 	77.17	 & 42.17	 & 76.51	 & 34.57 \\
\hline
CLIP-E Contrastive [SC, SSC] & 83.12* & 81.49* & 53.70* & 80.40 & \textbf{44.94} & 76.64 & 35.66 \\
\hline

\end{tabular}
\label{tab:crossdataset_results}
\end{center}
\end{table*}

We test both CLIP-E models approaches and CLIP Zero shot on 4 different datasets (7 benchmarks). The results are shown in Table \ref{tab:crossdataset_results}. We use 'X' to denote when an experiment can not be computed. Up to our knowledge, the only work that includes cross-dataset evaluation is Panda et al. \cite{panda2018contemplating} in the binary classification setup. 

For the binary classification for small datasets like IAPS, Emotion-6 and EmotionROI, CLIP-E Cross-Entropy produces better accuracies than CLIP-E Contrastive. 
%This is because the CLIP-E Cross-Entropy has been trained to learn the specific task of classifying between positive vs negative sentiment images, while in CLIP-E Contrastive the training for positive vs. negative sentiment is softer. 
However, the CLIP-E Contrastive obtains accuracies that are slightly higher than the CLIP Zero-Shot. 
Regarding the Pandas et al. \cite{panda2018contemplating} results, the authors trained a customized ResNet with curriculum learning for the WEBEmo dataset. However, when they tested in other datasets, they achieve accuracies that are equivalent to CLIP Zero-Shot for Emotion-6. The CLIP Zero-Shot is better for FI dataset than Pandas et al. \cite{panda2018contemplating}. That experimentally illustrates the potential of large Vision-Language models in generalizing.

For the fine grained classification of Emotion-6, we could test CLIP-E Cross-Entropy with 6 classes because the categories are the same of WEBEmo's. However, notice that the emotion taxonomies of each dataset could have been different. So, although CLIP-E Cross-Entropy reports a remarkable accuracy (58.39), if the Emotion-6 taxonomy were different than the WEBEmo taxonomy, it would have been not possible to test directly test CLIP-E Cross-Entropy (trained on WEBEmo) on Emotion-6. 

For the case of the large scale dataset FI, the best performance is achieved by CLIP Zero-Shot, surpassing the ResNet of Panda et al. \cite{panda2018contemplating} and the CLIP-E approaches. The reason could be that FI is a dataset that contains different kinds of images than WEBEmo. FI is composed of images that illustrate natural scenes and situations that evoke sentiments (that are probably similar to the images that CLIP has been  trained), while WEBEmo contains images of people and symbols, where the sentiments are represented by facial and body gestures, or conceptual meanings behind symbols or objects. However, it is important to notice that CLIP-E accuracies are higher than Panda et al. \cite{panda2018contemplating}.
%and the CLIP-E Contrastive seems to improved at large scale manner when all the image caption and description information is available.
 
%We can not compare our accuracies obtained in models that are trained in this specific dataset task and taxonomy.

Notice that CLIP-E Contrastive has more testing versatility than cross-entropy approaches. In the CLIP-E Cross-Entropy and other classic classification models, the only way to leverage and test the knowledge learnt is when the other dataset has the same taxonomy, or finetuning the model using the taxonomy of the particular dataset (which requires re-training).
%, but with the later, the model can learn the biases and particularities of the new dataset. 
Thus, CLIP-E Cross-Entropy can only be tested in other datasets in the binary category set up, and with Emotion-6 in the 6 categories because WEBEmo has the same taxonomy.
%however this particular case it is not always possible.
In contrast, CLIP-E Contrastive offers the flexibility to be tested on any taxonomy, due to its inference formulation, based on an arbitrary collection of prompts and the cosine similarity. Thus, while CLIP-E Contrastive is not obtaining the highest accuracies, it has the versatility to be tested on any dataset. Furthermore, although CLIP-E Contrastive does not systematically obtain the highest accuracies, it obtains accuracies comparable with other approaches, and obtains the second best results in 3 out of the 7 tested benchmarks.

\section{Discussion}
\label{sec:discussion}

\begin{figure*}[h]
\centering
\includegraphics[width=\textwidth]{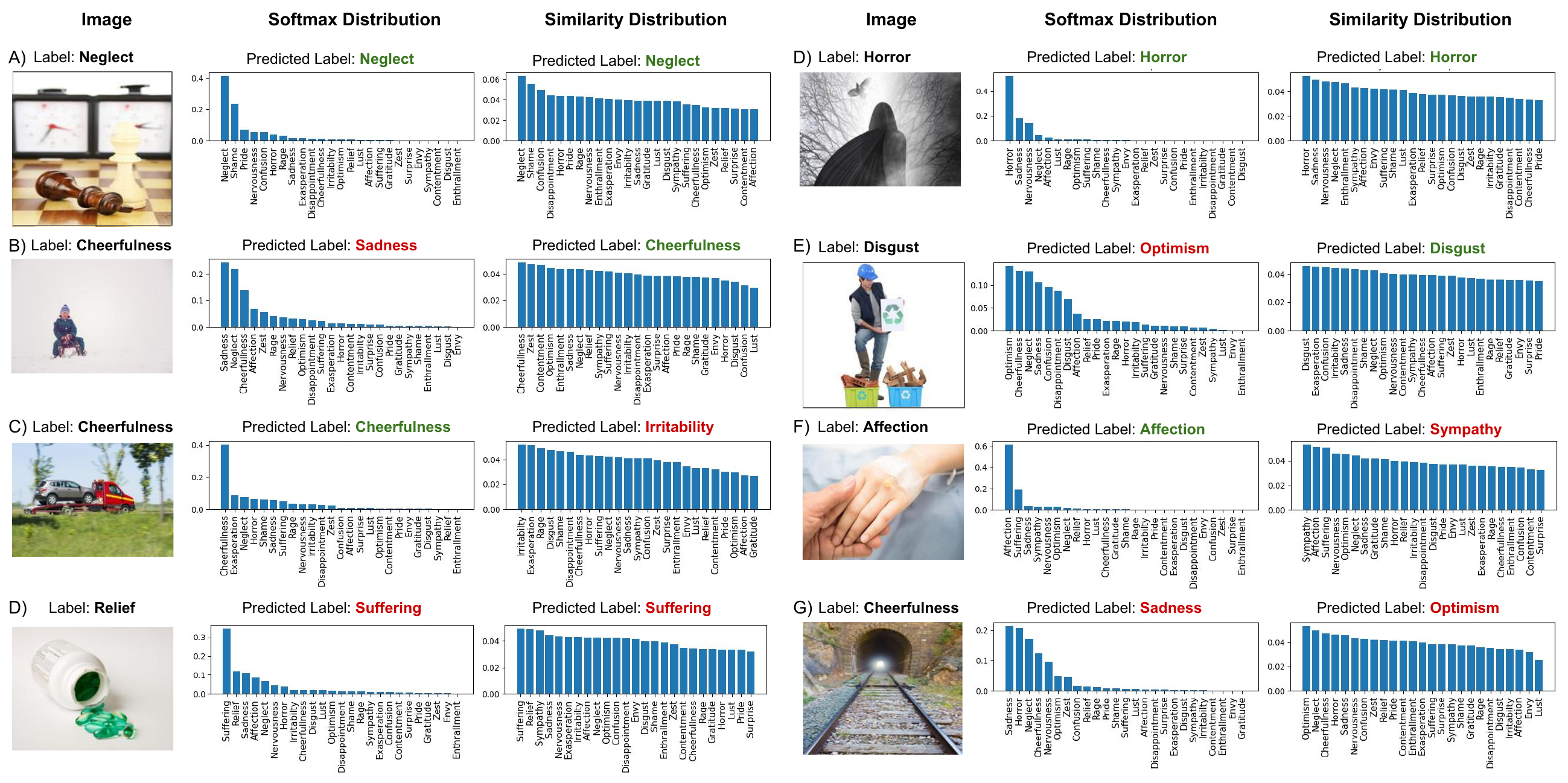}
\caption{Qualitative results of CLIP-E Cross-Entropy (Softmax Distribution provided) and CLIP-E Contrastive (Similarity Distribution provided) on WEBEmo test images. The Distribution graphs show the categories in decreasing order according probability (in the case of Softmax Distribution) and similarity (in the case of Similarity Distribution).}
\label{fig:results_qualitative}
\end{figure*}

\subsection{Multiple affective interpretations of an input}

Fig.\ref{fig:results_qualitative} shows qualitative results for CLIP-E Cross-Entropy (we show per each image the softmax distribution, sorting the emotion categories by probability, in decreasing order) and CLIP-E Contrastive (we show the normalized similarity distribution, sorting the categories by similarity, in decreasing order) trained and tested on WEBEmo. 

Usually, Visual Sentiment Analysis is approached as a multi-class classification problem, which assumes there is an unique ground truth category per image. However, while observing the qualitative results, it is reasonable to question this approach. For example, in Fig.\ref{fig:results_qualitative}.B, CLIP-E Contrastive predicts the ground truth label \textit{Cheerfulness}. One possible explanation for this output is the semantic content of the image, which is more related to a positive and fun situation (we noted that the IC of this  image is 'a photo that contains a kid sledding in snow at the mountain', which is a description of a positive experience). In contrast, CLIP-E Cross-Entropy classifies the image as with \textit{sadness}, and this seems also reasonable, since the image can also evoke loneliness in a desolated and grayish environment. Another example is Fig.\ref{fig:results_qualitative}.E, where the semantic information is not lost and concepts like 'recycling' are linked to 'garbage' and indirectly closer to \textit{disgust}, while the output of CLIP-E Cross-Entropy classifies this picture of a standing man as \textit{optimism} (by manually exploring the WEBEmo dataset we found several images of confident standing men labelled with positive emotion categories, which suggests that CLIP-E trained with Cross-Entropy loss was able of learning this visual pattern). Another interesting observation can be made around the idea of multiple possible interpretations of the same image. For example, Fig.\ref{fig:results_qualitative}.D is labeled as \textit{relief}, while both CLIP-E Cross-Entropy and Contrastive produce the output \textit{suffering} (which is an emotional state that occurs before relief). 
%Similarly, Fig.\ref{fig:results_qualitative}.F is labeled as \textit{affection}, which is also the output of CLIP-E Cross-Entropy. However, CLIP-E Contrastive outputs \textit{sympathy}, which is also a reasonable interpretation of this image showing a hand holding an injured hand. Notice that the second category for CLIP-E Cross-Entropy, and the third for CLIP-E Contrastive is \textit{suffering}, which again is a reasonable interpretation for an image that contains an injured hand. 
Since humans can make different interpretations of the same image, one might think that automatic Visual Sentiment Analysis systems  should be able of doing the same, and for that we need new benchmarks that include multiple interpretations of the same sample, as well as evaluation metrics to quantify how models perform on the multiple interpretation task. 

These observations suggest questions that we should consider in future work on Visual Sentiment Analysis, including data collection efforts, where we could consider labelling images with multiple categories (multi-label) instead of labelling them with a single category (multi-class). While multi-class classification is also the most common apparent emotion recognition, there are recent works that are starting to approach the task with a multi-label perspective (e.g. \cite{kosti2019context}).

\subsection{Dataset bias and the need for cross-dataset evaluation}

%First, we observe that the softmax distribution from the CLIP-E Cross-Entropy has a 'long tail' distribution, while the cosine similarity  distribution is more uniform. 
Fig.\ref{fig:results_qualitative} also illustrates the diversity of the WEBEmo images: some of them are more object centric (e.g. Fig.\ref{fig:results_qualitative}.A or Fig.\ref{fig:results_qualitative}.D), while others are more scene centric (Fig.\ref{fig:results_qualitative}.C or Fig.\ref{fig:results_qualitative}.G). Some of them seem natural scenes (e.g. Fig.\ref{fig:results_qualitative}.G), others are posed pictures (e.g Fig.\ref{fig:results_qualitative}.B), and others have an associated conceptual message (e.g. Fig.\ref{fig:results_qualitative}.A, which symbolizes winning/losing). While WEBEmo is a large and diverse dataset, we also observed that each of the datasets considered in this work have their own particularities and biases. This is the reason why we face the specialization vs. generalization challenge: the better a model performs on WEBEmo, at some point it starts performing poorly with other datasets.  Concretely, after training CLIP-E with WEBEmo, we tested the models obtained after each training epoch on the other benchmark datasets. Overall, the CLIP-E Contrastive with WEBEmo needed 6 epochs to converge and CLIP-E Cross-Entropy needed 5. However, the best performance on most of the other datasets was obtained by the first or second epoch model for both CLIP-E Contrastive and CLIP-E Cross-Entropy. These results show that as the model specializes on WEBEmo, it loses the ability to generalize on other datasets. These observations motivate the importance of cross-dataset evaluation when testing Visual Sentiment Analysis models, since it is a mechanism to evaluate generalization capacity. Furthermore, the results of our cross-dataset evaluation clearly show the capacity of CLIP Zero-Shot and CLIP-E approaches to generalize across different datasets, when compared to models specifically designed and trained just for the Visual Sentiment Analysis task. Under the current Computer Vision shift to large Vision-Language foundation models, we might want to put more effort on understanding these models and their use for Visual Sentiment Analysis and other affect recognition tasks.

\section{Conclusions}
We present a study on leveraging knowledge encoded in the CLIP embedding space for the task of Visual Sentiment Analysis. We experiment with two simple architectures built on top of the CLIP embedding space with Cross-Entropy loss (CLIP-E Cross-Entropy) and Contrastive loss (CLIP-E Contrastive), respectively. The CLIP-E architectures are trained on WEBEmo dataset, on top of freezed CLIP embeddings. Our results show that these architectures are able of obtain accuracies competitive or even superior in fine grained categorization on the WEBEmo benchmark, when compared to SOTA models. Additionally, our cross-dataset evaluations show that the CLIP-E architectures, which strongly rely on the knowledge of CLIP, are able to generalize better across different datasets. Our results motivate interesting discussions, such as the need for new Visual Sentiment Analysis benchmarks containing multiple interpretations of the same image, the interest of cross-dataset evaluation for testing the generalization capacity of the models, and the motivation for further exploring the use of large vision-language models for visual affect recognition tasks more generally. 
%,  \cb{Overall, our work demonstrates the potential of the large vision-language models like CLIP for advancing research in Visual Sentiment Analysis. However, our work also highlights potential limitations of current approaches, such as the common multi-class approach and the generalization capacity of customized deep learning architectures. These findings motivate further exploration of multi-label approaches and leveraging knowledge from vision-language models for Visual Sentiment Analysis and other affect recognition tasks.que tal asi?}

%%  ---- START: Added 2022-03-15 for ACII2022 onwards -----
\section*{Ethical Impact Statement}
Visual Sentiment Analysis is a challenging problem. While CLIP-E is obtaining state-of-the-art accuracies on fine grained categories in WEBEmo, we can not claim that CLIP-E (or any existing methods) can accurately recognize affect in images. There is a lot of research to be done yet to have automatic systems able to richly and diversely recognize affective content in images as humans would do.
%correctly and without biases. 
Models like CLIP-E are trained on biased datasets, and the bias is present both in the image instances (most of the images were acquired in western countries, and there are images that show negative biases towards certain groups --e.g. we observed that images containing women are more frequent labeled to negative emotions like \textit{envy} than images containing men) and in the annotations (most of the images are labeled by a limited number of annotators and the labels do not represent the diversity of opinions in terms of affect perception). 
The WEBEmo dataset is composed of images crawled from the internet. CLIP original approach has been trained with images and text captions crawled from the internet. It is known that these types of data collections and models contain biases, some of which reflect the biases embedded in human society.
While the purpose of this work is to advance research on Visual Sentiment Analysis, the outcome of this work could potentially be used in practical real cases. For example for estimating the emotional impact of images in advertisement, or the emotional impact of movie clips, or for retrieving images with emotional content. Any practical use case of Visual Sentiment Analysis models should be properly regulated. Furthermore, if the use might directly impact a person, the person should give their consent for the use of such models. More generally, any use in real scenarios of Visual Sentiment Analysis models has to be validated for the corresponding ethics committees. Additionally, more effort needs to be made to provide explainability capacity to the models, to be able of understanding their behavior. Finally, regarding environmental matters, even though our model leverages on CLIP knowledge and requires minimal effort to be trained, original large scale vision-language models are computationally expensive and have a negative impact on carbon fingerprint.
%However, these are very important and interesting topics of research in the context of visual sentiment analysis. .

\section*{Acknowledgments}

This work is partially supported by the Spanish Ministry of Science, Innovation and Universities RTI2018-095232-B-C22 to A.L; a Google unrestricted gift to A.L; and UOC PhD Grant to C.B. We also thank NVIDIA for their hardware donations.

%%  ---- END: Added 2022-03-15 for ACII2022 onwards -----

\bibliographystyle{IEEEtran}
\bibliography{IEEEabrv,bibliography}

\end{document}